\documentclass{article} 
\usepackage[preprint]{colm2026_conference}
\usepackage{microtype}
\usepackage{hyperref}
\usepackage{url}
\usepackage{booktabs}
\usepackage{multirow}
\usepackage{graphicx}
\usepackage{float}
\usepackage{rotating}
\usepackage[most]{tcolorbox}
\usepackage[skip=3pt]{caption}
\usepackage{enumitem}
\usepackage{tabularx}
\usepackage{ragged2e}
\usepackage{adjustbox}
\usepackage{bm}
\usepackage{wrapfig}
\usepackage{lineno}

\definecolor{darkblue}{rgb}{0, 0, 0.5}
\hypersetup{colorlinks=true, citecolor=darkblue, linkcolor=darkblue, urlcolor=darkblue}

\title{Where Larger Models Excel:\\
The Primacy of Constraint-Guided Reasoning}

\author{Guan-Yi Lin\thanks{Code: \url{https://github.com/lgyeee/Where_Do_Larger_Model_Excel}} \\
  National Chengchi University \\
  Taipei, Taiwan \\
  \texttt{111703052@g.nccu.edu.tw}
  \And
  Hen-Hsen Huang \\
  Academia Sinica \\
  Taipei, Taiwan \\
  \texttt{hhhuang@iis.sinica.edu.tw}
}

\begin{document}

\ifcolmsubmission
\linenumbers
\fi

\maketitle

\begin{abstract}
Larger language models consistently outperform smaller ones on reasoning benchmarks, yet the reasoning differences underlying this gap remain underexplored. Across benchmarks in mathematics, physics, chemistry, and programming, we observe stable performance gaps: averaged over datasets, Qwen3-32B outperforms Qwen3-8B by 6.43\%, while GPT-OSS-120B exceeds GPT-OSS-20B by 7.38\%. To study the reasoning differences behind these gains, we develop \textbf{AdvCluster}, an automated framework that identifies questions where the larger model shows a stable advantage, extracts fine-grained advantage descriptions from paired reasoning traces produced by larger and smaller models, and organizes them through semantic clustering with quantitative evaluation and selection guided by a reviewer model. Our analysis yields a systematic taxonomy of larger model reasoning advantages, spanning both common advantages that recur across domains and specialized advantages associated with particular domains. Across these patterns, a recurring theme is \textit{Constraint-Guided Reasoning}: larger models are better at identifying explicit and implicit constraints, organizing them into structured reasoning, and using them to rule out infeasible paths and verify intermediate steps.
\end{abstract}

\section{Introduction}
While larger language models consistently outperform smaller ones on complex reasoning benchmarks, the reasoning processes underlying this advantage remain insufficiently understood. Existing evaluations focus primarily on aggregate outcomes such as accuracy, which reveal little about where larger models excel or how their reasoning differs from that of smaller models\citep{DBLP:conf/iclr/LightmanKBEBLLS24}.

Prior works primarily focus on enhancing smaller models via distillation to address reasoning bottlenecks \citep{DBLP:conf/acl/HsiehLYNFRKLP23, DBLP:journals/corr/abs-2311-11045}. While some evaluate reasoning through process supervision \citep{DBLP:conf/iclr/LightmanKBEBLLS24} or holistic benchmarks \citep{liang2023holistic}, the field still lacks a systematic framework for empirically analyzing the qualitative reasoning advantages of larger models. 

\begin{figure}[]
    \centering
    \makebox[\textwidth][c]{\includegraphics[width=1\textwidth, trim=0 10pt 0 0, clip]{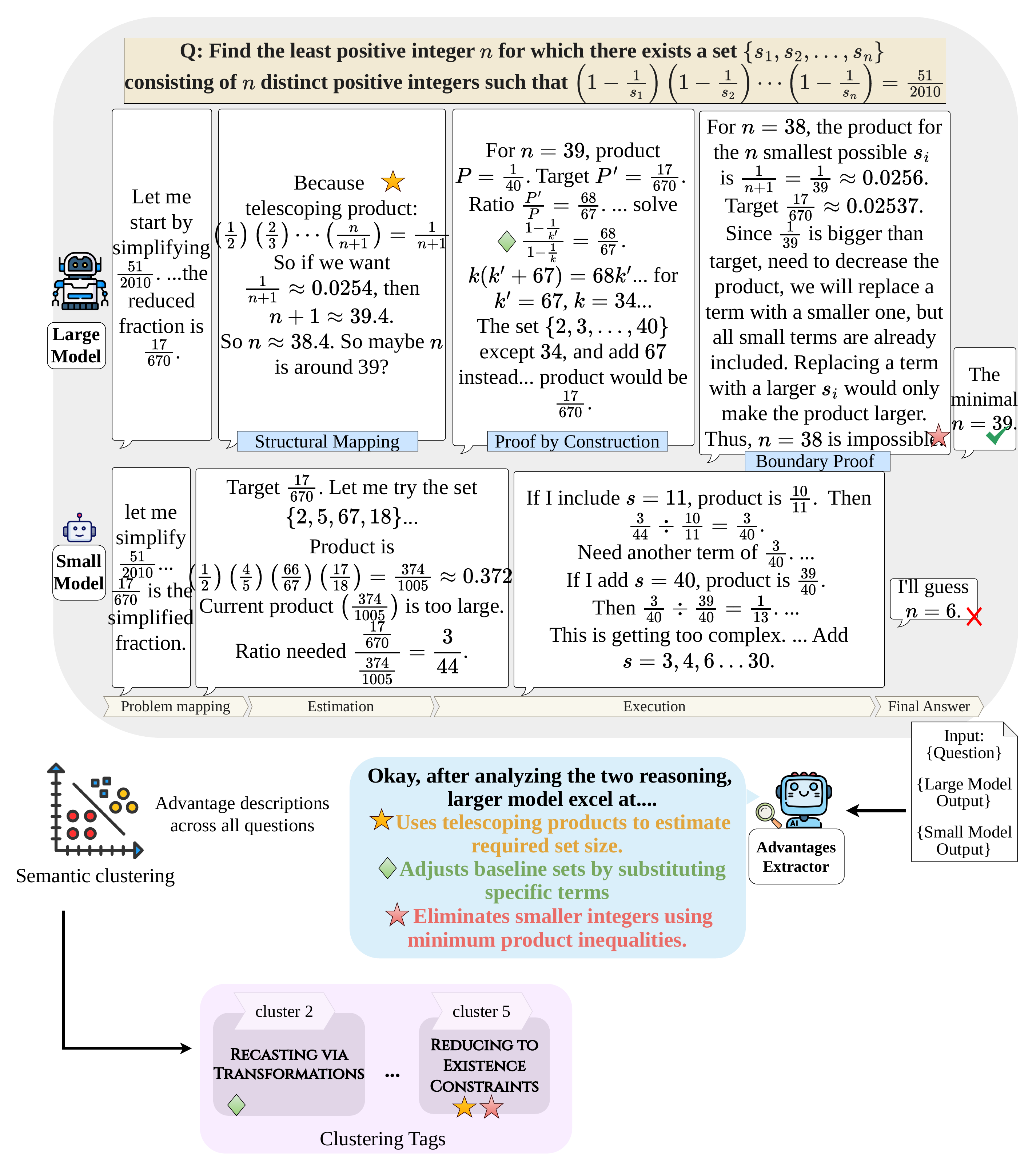}}
    \vspace{2pt}
    \caption{Illustration of the \textbf{AdvCluster} framework, exemplified by a comparative analysis of Qwen3-32B and Qwen3-8B on a mathematics problem. The upper and lower panels contrast the reasoning traces of the larger and smaller models, with annotated symbols ($\star, \diamond$) highlighting specific steps where the larger model demonstrates strategic advantages. An LLM-based advantage extractor converts these local differences into textual advantage descriptions. Across many such questions, these advantages are aggregated via semantic clustering into cluster tags—such as \textsc{Recasting via Transformations} and \textsc{Reducing to Existence Constraints}—that characterize the broader \textit{Constraint-Guided Reasoning} pattern. }
    \label{fig:qwen3case}
\end{figure}

In this work, we present a cross-scale analysis of reasoning, specifically comparing larger models against their smaller counterparts within the same model family. Inspired by \citet{DBLP:conf/emnlp/YinSHQZ25}, we develop \textbf{AdvCluster}, an automated advantage discovery framework that first identifies benchmark questions where the larger model consistently outperforms the smaller model, and then compares their reasoning traces to extract advantage descriptions. We then organize these descriptions into clusters through a semantic processing pipeline, and use quantitative metrics together with a reviewer model to select among candidate clustering solutions based on its consistency, distinctness, and granularity. Figure~\ref{fig:qwen3case} illustrates the AdvCluster framework on a representative math example, showing how the reasoning advantages of larger models emerge naturally from the data.

Our analysis categorizes these reasoning advantages into two types: \textbf{common advantages}, which recur across multiple domains, and \textbf{specialized advantages}, which are tied to domain-specific knowledge. A primary common pattern identified by AdvCluster across various domains is \textit{Constraint-Guided Reasoning}. As illustrated in Figure~\ref{fig:qwen3case}, while smaller models often rely on unguided trial and error, larger models tend to reformulate problems using explicit constraints to systematically guide the solution process.

\section{Motivation and Preliminary Observations}
Chain-of-thought prompting improves performance on complex reasoning tasks while exposing intermediate reasoning steps, making reasoning traces analyzable \citep{DBLP:conf/nips/Wei0SBIXCLZ22, DBLP:conf/iclr/0002WSLCNCZ23}.
This increased transparency has motivated a growing body of work on reasoning trajectory analysis~\citep{DBLP:conf/naacl/YeoSGC24, DBLP:conf/acl/WangM0S0Z023}.

\textbf{Larger models generally perform better.}
A separate line of work shows that language model capability generally improves with scale. \citet{DBLP:journals/corr/abs-2001-08361} characterize these gains through neural scaling laws over parameters, data, and compute. Subsequent studies further suggest that some complex capabilities, including multi-step reasoning, emerge only at sufficiently large scales. \citet{DBLP:journals/tmlr/WeiTBRZBYBZMCHVLDF22} describe such behaviors as emergent abilities, namely capabilities absent in smaller models but present in larger ones.

Together, these findings indicate that larger models tend to achieve stronger reasoning performance than their smaller counterparts. However, a central question remains: where do larger models exhibit consistent reasoning advantages over smaller ones?

Most existing work instead focuses on improving smaller models, for example by addressing specific bottlenecks such as reasoning data quality through distillation~\citep{DBLP:conf/coling/ZhaoZZ24}.
While such efforts may indirectly suggest where larger models hold advantages, they do not provide a systematic empirical account of how reasoning behavior differs between larger and smaller models.

\textbf{Preliminary results.}
We first conduct a preliminary experiment across four domains: mathematics, physics, chemistry, and programming.
Our evaluation covers a diverse set of reasoning benchmarks, including HHMT~\citep{DBLP:journals/corr/abs-2505-23281}, Omni-MATH~\citep{DBLP:conf/iclr/Gao0YCMDLMCXTWZ25}, JEEBench~\citep{DBLP:conf/emnlp/AroraSM23}, OlympiadBench~\citep{DBLP:conf/acl/2024-1}, GPQA~\citep{DBLP:journals/corr/abs-2311-12022}, and CRUXEval~\citep{DBLP:conf/acl/XuC00LHHC025}. 
We evaluate two reasoning model families at different scales: Qwen3-8B vs.\ \ Qwen3-32B, and GPT-OSS-20B vs.\ GPT-OSS-120B \citep{DBLP:journals/corr/abs-2505-09388,DBLP:journals/corr/abs-2508-10925}.

Our preliminary results show a consistent improvement from smaller models to their larger counterparts on these reasoning benchmarks. 
On average across all evaluated datasets, Qwen3-32B outperforms Qwen3-8B by 6.43\%, while GPT-OSS-120B outperforms GPT-OSS-20B by 7.38\%. These preliminary results confirm the performance gap between larger and smaller models in our setting and motivate a closer investigation into where these gains arise in the reasoning process.
Detailed results per data set are provided in Appendix~\ref{detailed_acc}.
\begin{table}[H]
\centering
\begin{tabular}{lcc}
\toprule
\textbf{Model Family} & \textbf{Model} & \textbf{Accuracy} \\
\midrule
\multirow{2}{*}{Qwen3} 
& 8B  & \(59.35 \pm 5.34\) \\
& 32B & \(\mathbf{65.78 \pm 5.13}\) \\
\midrule
\multirow{2}{*}{GPT-OSS} 
& 20B  & \(59.51 \pm 2.10\) \\
& 120B & \(\mathbf{66.89 \pm 1.73}\) \\
\bottomrule
\end{tabular}
\caption{Average accuracy (\%) across all evaluated reasoning benchmarks.}
\label{tab:preliminary-results}
\end{table}

\section{Methodology}
\subsection{Dynamic Advantage Classification}
Reasoning advantages are difficult to categorize using a predefined taxonomy, since the forms of model superiority vary across tasks and domains. This makes static category assignment by an LLM judge restrictive and potentially unreliable. \citet{DBLP:conf/emnlp/YinSHQZ25} address a related challenge in mathematical error analysis with a dynamically adaptive framework, in which categories are induced from data rather than specified in advance. Inspired by this idea, we develop a data-driven pipeline for reasoning advantage analysis. Instead of imposing a fixed taxonomy, we first extract fine-grained advantage descriptions from empirical comparisons between larger and smaller models' reasoning traces, and then organize these descriptions through semantic clustering. This allows recurring reasoning advantage categories to emerge naturally from the data.

\subsection{Our Framework: AdvCluster}
AdvCluster consists of three stages: Analysis Question Set, Advantage Extraction, and Semantic Clustering. In the first stage, we construct an analysis question set by identifying benchmark questions on which the larger model consistently outperforms the smaller model across repeated runs. In the second stage, we construct paired comparisons between the larger and smaller models' reasoning traces for each question in the analysis set, and use a LLM as an Advantages Extractor to produce fine-grained advantage descriptions from these comparisons. In the third stage, we embed and cluster the extracted descriptions to induce an interpretable taxonomy of larger model reasoning strengths.

\subsubsection{Analysis Question Set}
\textbf{Gap-Based Filtering.}
Because our goal is to compare reasoning differences between larger and smaller models, we first identify questions for which the larger model consistently outperforms the smaller one across repeated trials. This step isolates analysis questions that reflect stable performance differences.

For each question $q$, we evaluate both models over $T$ independent trials. Let $c_M(q,t)=1$ if model $M$ answers $q$ correctly on trial $t$, and $0$ otherwise. We define the pass rate of model $M$ on $q$ as
$$
\mathrm{PassRate}_M(q)=\frac{1}{T}\sum_{t=1}^{T} c_M(q,t),
$$
i.e., the proportion of trials on which $M$ answers $q$ correctly.

We define the performance gap for question $q$ as
$$
\Delta(q)=\mathrm{PassRate}_{M_L}(q)-\mathrm{PassRate}_{M_S}(q),
$$
where $M_L$ and $M_S$ denote the larger and smaller models, respectively. This quantity measures how much more often the larger model answers question $q$ correctly than the smaller model across repeated trials.

We retain questions with sufficiently large $\Delta(q)$ to form the analysis question set. The next stage constructs reasoning comparisons only on this filtered set.

\subsubsection{Advantage Extraction}
For each question $q$ in the analysis question set, we construct reasoning comparisons between the larger and smaller models. Each comparison instance consists of a pair of reasoning traces generated on the same question, one from the larger model and one from the smaller model.

Let $r^{L}_{q,i}$ and $r^{S}_{q,i}$ denote the larger model and smaller model reasoning traces in the $i$-th comparison instance for question $q$, respectively. We then use an Advantages Extractor $\mathcal{J}$ to compare the paired traces and produce a set of advantage descriptions:
$$
Z_{q,i}=\mathcal{J}(q, r^{L}_{q,i}, r^{S}_{q,i}),
$$
where each element $z \in Z_{q,i}$ is a natural-language description of a reasoning advantage exhibited by the larger model over the smaller model in that comparison. In this way, each reasoning pair generates  a small set of advantage descriptions.

Concretely, the extractor is implemented with the following prompt:
\begin{tcolorbox}[colback=white, colframe=black!50, boxrule=0.1pt, title=Prompt]
You are an advantage extraction expert. \\
Model\_A is correct; Model\_B is incorrect. \\
TASK: \\
Compare their reasoning \dots \\
\dots extract 2--5 advantage objects explaining why Model\_A succeeds. \\
Problem: \{q\} \\
Model\_A reasoning: \{larger\_model\_reasoning\} \\
Model\_B reasoning: \{smaller\_model\_reasoning\} \\
...
\end{tcolorbox}
The complete prompt is provided in Appendix~\ref{llmanalyzeprompt}

\subsubsection{Semantic Clustering}
\label{3.2.3Semantic Clustering}

After advantage extraction, we obtain a collection of natural-language advantage descriptions from all comparison instances constructed on the analysis question set. These descriptions summarize how the larger model outperforms the smaller model across questions from different domains. Let
$$
\mathcal{Z}=\{z\}
$$
denote the set of all extracted advantage descriptions.

\textbf{Advantage Encoding and Preprocessing.} We encode each description $z \in \mathcal{Z}$ into an embedding vector
$$
e=\phi(z)\in\mathbb{R}^{d},
$$where $\phi$ denotes the embedding model. 

As multiple comparison instances for the same question may yield highly similar advantage descriptions, directly clustering all descriptions can bias cluster centroids toward duplicated patterns. We therefore perform deduplication in the embedding space using a greedy procedure with a cosine-similarity threshold of $0.95$ \citep{DBLP:conf/naacl/GuptaRPPUW25, DBLP:conf/lrec/GyawaliAK20}.

We then reduce the embedding dimensionality via PCA before clustering to mitigate the difficulty of clustering in high-dimensional vector spaces, following common embedding-based text clustering pipelines \citep{DBLP:journals/corr/abs-2203-05794,eklund-etal-2023-empirical}.
$$
\tilde{e}=\psi(e)\in\mathbb{R}^{d'}, \qquad d' < d.
$$
Let
$$
\mathcal{E}=\{\tilde{e}\}
$$
denote the deduplicated and low-dimensional advantage embedding vectors.

\textbf{Clustering and Candidate Generation.}
We apply K-means clustering to $\mathcal{E}$ under multiple candidate settings $(d', K)$, where $d'$ is the PCA dimension and $K$ is the number of clusters. For each fixed $d'$, we sweep over a range of $K$ values and quantitatively evaluate the resulting candidate clustering solutions using Davies--Bouldin Index (DBI) as the primary criterion and the Silhouette score as a secondary reference. Based on this quantitative evaluation, we retain a smaller set of quantitatively favorable candidate clustering solutions.

For each retained candidate clustering solution, we map the low-dimensional embeddings assigned to each cluster back to their underlying natural-language advantage descriptions. We then use a summarization model to produce, for each cluster:
\begin{itemize}[leftmargin=1.5em, topsep=-2pt, partopsep=0pt, itemsep=0pt, parsep=0pt]
    \item a short \textbf{tag};
    \item a concise \textbf{definition} of the shared reasoning advantage pattern.
\end{itemize}
The detailed prompt for summarization model is provided in Appendix~\ref{llmsummrization}.

\textbf{Final Clustering Selection.}
The quantitative evaluation above narrows the options to a smaller set of retained candidate clustering solutions. However, for downstream results analysis, we seek a clustering solution that is not only quantitatively supported but also semantically well separated, interpretable, and appropriately granular. We therefore use a reviewer model to compare the retained candidate clustering solutions and assess which one is most suitable overall, based on criteria such as internal coherence, separation between clusters, semantic interpretability, and granularity. The prompt for reviewer model is provided in Appendix~\ref{llmreviewer}.

Finally, based on both the quantitative metrics and the reviewer model's assessment, we select the clustering solution used for downstream results analysis. The selected clustering solution determines the final number of clusters $K$ and yields
$$
\{(C_k, t_k, d_k)\}_{k=1}^{K},
$$
where $C_k$ is the set of advantage descriptions assigned to the $k$-th cluster, $t_k$ is its tag, and $d_k$ is its definition.
\section{Experiment}
\subsection{Setup}
\paragraph{Models and Benchmarks.}
We study two larger--smaller model pairs from the same family: Qwen3-32B vs.\ Qwen3-8B, and GPT-OSS-120B vs.\ GPT-OSS-20B. We evaluate them on benchmarks from four domains: mathematics (HHMT, OMNI, JEEBench), physics (GPQA, JEEBench, Olympiad-level benchmarks), chemistry (GPQA, JEEBench), and programming (CRUXEval).

\paragraph{Repeated Inference and Analysis Question Set Construction.}
For each model on each dataset, we perform $T=10$ independent runs. For each question $q$, we compute the larger--smaller performance gap
$$
\Delta(q)=\mathrm{PassRate}_{M_L}(q)-\mathrm{PassRate}_{M_S}(q),
$$
and retain questions with $\Delta(q)\geq 0.6$. These retained questions form the analysis question set.

\subsection{Advantage Analysis Pipeline}

\paragraph{Advantage Extraction and Clustering.}
Applying the above filtering procedure yields analysis question sets of 115 questions for Qwen3 and 106 questions for GPT-OSS. For each question in the analysis question set, we construct paired reasoning comparisons between the larger and smaller models and use \texttt{Gemini 3 Pro} as the advantage extractor to generate advantage descriptions.

We encode the extracted advantage descriptions using OpenAI's \texttt{text-embedding-3-large}, then perform semantic deduplication and PCA before clustering the resulting embeddings under multiple candidate settings $(d', K)$. After deduplication, the final analysis corpus contains 1824 advantage descriptions for Qwen3 and 1963 for GPT-OSS.

For each retained candidate clustering solution, we use \texttt{gpt-5.2} as the summarization model to generate cluster tags and definitions. We then use \texttt{gpt-5.2} as a reviewer model to assess candidate clustering solutions in terms of internal coherence, inter-cluster distinctness, interpretability, and granularity. Based on this assessment together with quantitative clustering metrics, we select the final clustering solution used in our results analysis.

Detailed domain level statistics and deduplication results are provided in Appendix \ref{Analysis Question Set by Subject}.

\section{Results and Analysis}

\begin{figure}[ht]
    \centering
    \includegraphics[width=0.7\linewidth]{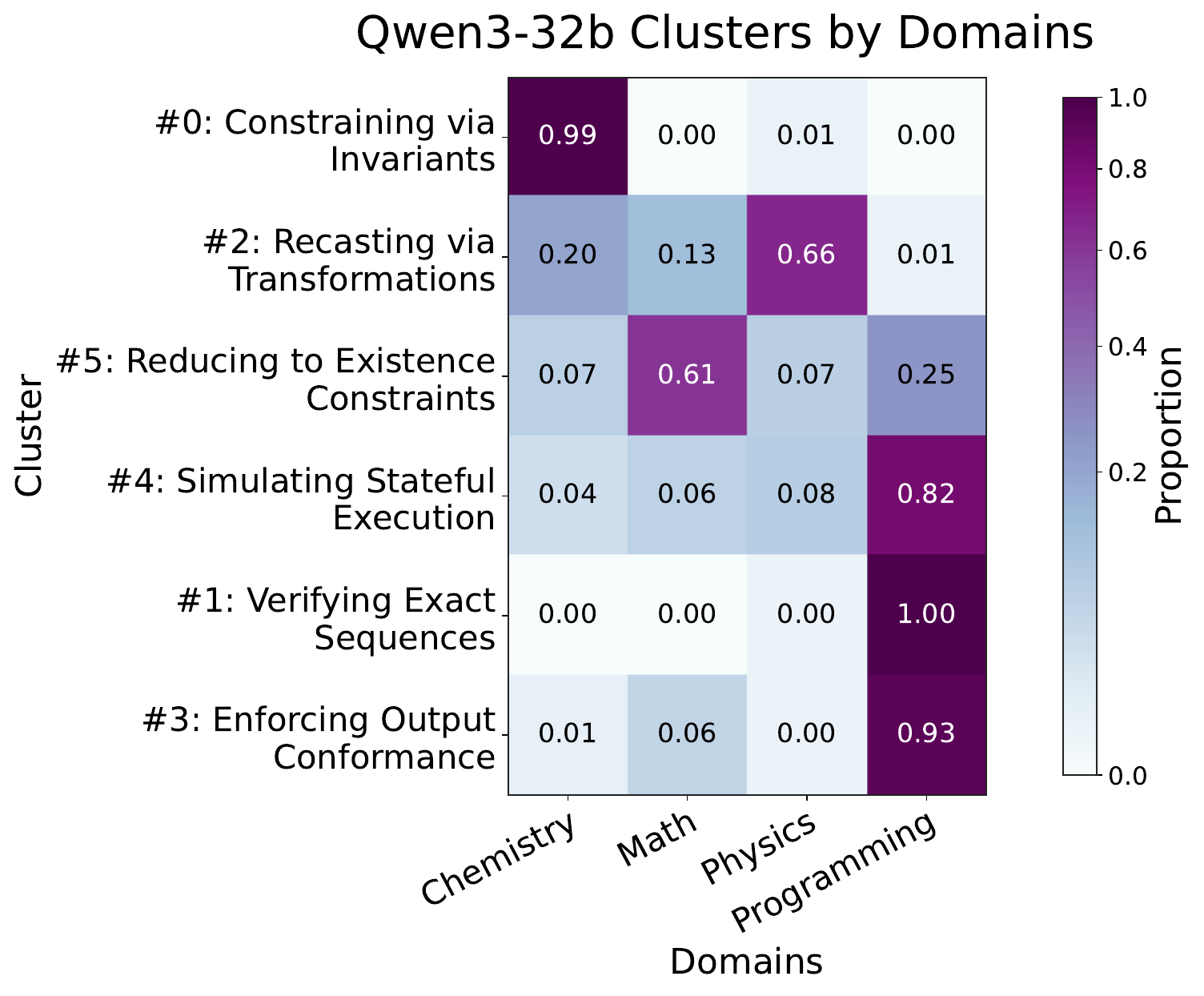}
    \vspace{-1pt} 
    
    \includegraphics[width=0.8\linewidth]{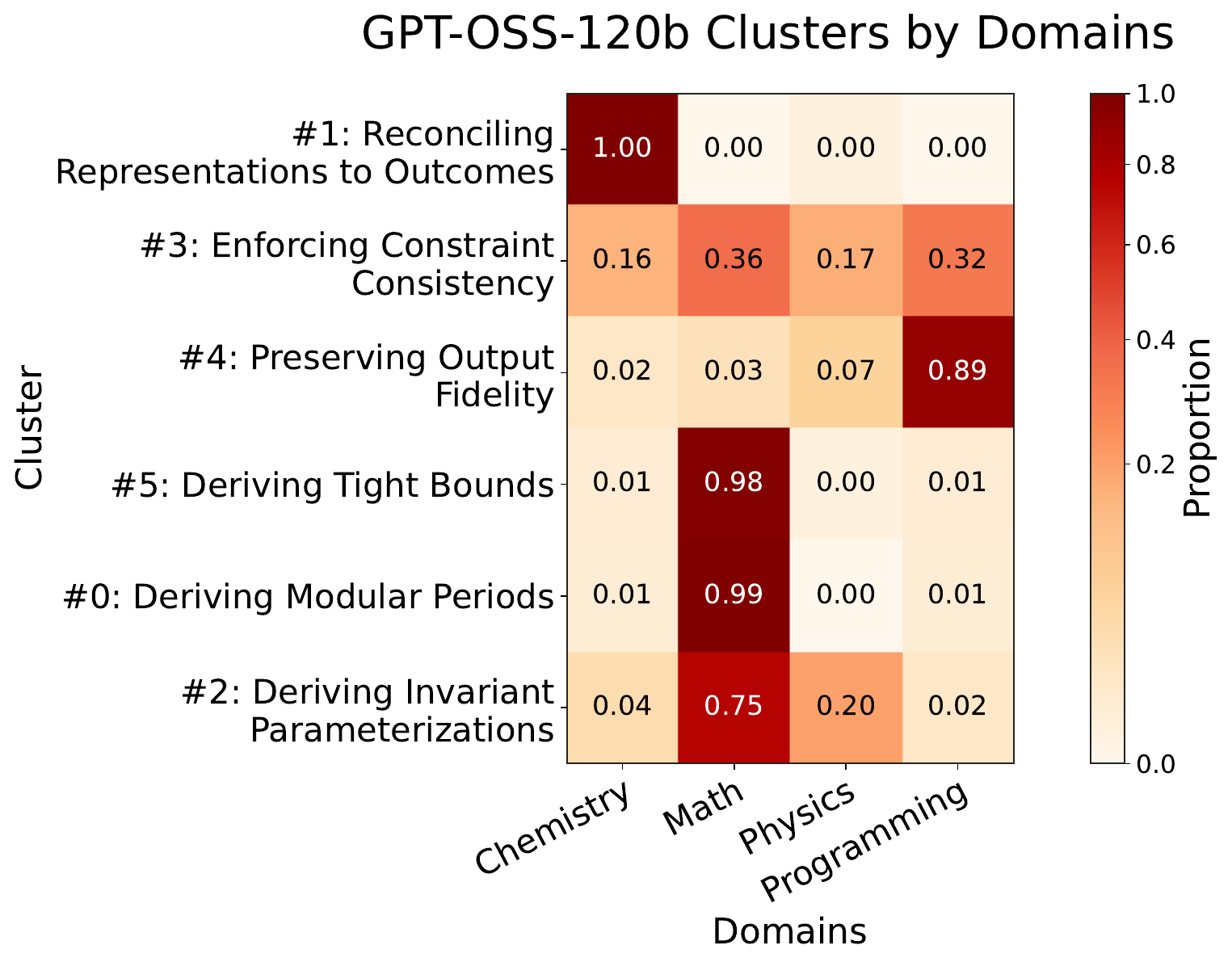}
    
    \caption{\textbf{Domain Distribution of Larger-Model Reasoning Advantages.}Heatmaps display the advantage clusters for Qwen3 (32B vs.\ 8B, top) and GPT-OSS (120B vs.\ 20B, bottom).  Cell values are row-normalized, so within each cluster (row), the proportions across domains sum to 100\%. The value in each cell shows the proportion of advantage descriptions in that cluster that come from a given domain. Cluster tags shown alongside the y-axis denote the semantic meaning of the clusters.}
    \label{fig:heatmap}
\end{figure}

On the basis of the quantitative clustering evaluation and reviewer model assessment described in Section~\ref{3.2.3Semantic Clustering}, we adopt the final clustering solutions of $d'=8$, $K=6$ for Qwen3 and $d'=4$, $K=6$ for GPT-OSS. These settings provide the best balance between quantitative cluster quality and semantic interpretability. The tags, definitions, and sizes of the final clusters are provided in Appendix~\ref{sec: Final Clustering Solutions}.

We then use heatmaps to examine how the resulting advantage clusters are distributed across domains. This analysis is motivated by the possibility that different domains may nonetheless share similar underlying reasoning patterns. As shown in Figure~\ref{fig:heatmap}, some clusters have substantial presence across multiple subjects, whereas others are concentrated in more specific domains. Based on these distributional patterns, we organize the reasoning advantages of larger models into two broad types:
\begin{itemize}
    \item \textbf{common advantages} recur across multiple subjects.
    \item \textbf{specialized advantages} are concentrated in a particular subject and are often tied to domain-specific representations or verification demands.

\end{itemize}
The former indicates that some reasoning strengths are shared  across tasks in various domain, while the latter is more evident in specific domains.

For clarity, the names shown in Figure~\ref{fig:heatmap} are the cluster tags derived from semantic clustering ~\ref{3.2.3Semantic Clustering}. We use these tags throughout the following discussion as the corresponding reasoning advantage clusters.

We further find that several clusters can be organized under a higher-level reasoning pattern: \textbf{Constraint-Guided Reasoning}. This refers to cases where the model structures the solution process around explicit or derived constraints, reformulates them into usable conditions, and uses them to restrict the search space, rule out invalid candidates, and support intermediate verification. We begin with this pattern, and then turn to additional reasoning advantages that are more specialized by subject.

\subsection{Constraint-Guided Reasoning}

The primary common advantages that emerged in our semantic clustering reveal a similar pattern: Constraint-Guided Reasoning. Rather than relying on trial and error over surface forms, larger models more often begin with a problem's explicit constraints and implicit constraints and reformulate them into more usable conditions, such as feasibility conditions, boundary conditions, invariant relations, and consistency checks. This allows the larger model to systematically narrow the solution space, remove invalid candidates, and verify intermediate steps in reasoning.

Qwen3 exhibits this pattern most clearly in its \textbf{common advantages} such as \textsc{\#5 Reducing to Existence Constraints} and \textsc{\#2 Recasting via Transformations}. 
Similar pattern appears in GPT-OSS through its \textbf{common advantages} like \textsc{\#3 Enforcing Constraint Consistency} and \textsc{\#2 Deriving Invariant Parameterizations}. Here, the larger model first organizes the problem into a unified set of constraints, then uses invariant relations or parameterized forms to derive exact formulas, inequalities, or bounds. The larger model uses them to maintain consistency with the original problem structure and to rule out conclusions that violate the constraints.

A representation case of Constraint-Guided reasoning is discussed earlier in Figure\ref{fig:qwen3case}. The larger model first performs structural reformulation, then advances the solution through boundary proof and reparameterization. By contrast, the smaller model stays closer to local trial and error over surface cues, making ineffective search and incorrect paths more likely.

This is also revealed in \textbf{specialized advantages}, where it is expressed through the solving tools of a particular domain:
\begin{itemize}
    \item In chemistry, Qwen3 exhibits \textsc{\#0 Constraining via Invariants}. Since chemistry reasoning process relies heavily on invariants--such as balances, conserved totals, configuration states, the model translates descriptions into a formal representation, derives quantities that must remain valid, and uses them to eliminate inconsistent candidates or revise conclusions when general heuristics conflict with harder constraints. 
    \item In mathematics, GPT-OSS naturally exhibits \textsc{\#0 Deriving Tight Bounds} and \textsc{\#5 Deriving Modular Periods}. The former advantage allows larger model to turn the problem into a counting system based on constraint and propagates local requirements into global feasibility or contradiction. In \textsc{\#5 Deriving Modular Periods}, it rewrites the problem as a modular system, tracks invariant residues, and derives periodic behavior from that constraint-guided reformulation.

\end{itemize}

\subsection{Other Reasoning Advantages}
\vspace{-5pt}
Beyond Constraint-Guided Reasoning, we observed other advantages emerge in specific domains:

\textbf{Execution Tracking.} 
This is primarily seen in programming tasks. Qwen3 demonstrates this in \textsc{\#4 Simulating Stateful Execution.} Larger models track operations step-by-step and update state variables after every single action. Throughout this process, strict control flow is maintained. Finally, intermediate cross-checks effectively eliminate state drift and missed updates. 

\textbf{Format Control.} 
This reasoning advantage appears in programming tasks. For instance, in Qwen3's \textsc{\#1 Verifying Exact Sequences}, the model precisely tracks elements and boundaries; consequently, it verifies the constructed output against properties like total length and delimiter placement. Similarly, in \textsc{\#3 Enforcing Output Conformance}, it identifies structural and syntax constraints early on, then seamlessly converts intermediate results into the required format. Furthermore, GPT-OSS exhibits a related pattern in \textsc{\#4 Preserving Output Fidelity}. During complex operations, this model maintains  integrity at character level and corrects minor local anomalies. Ultimately, these clusters demonstrate that larger models excel at presenting reasoning in the exact required form.

\textbf{Representation Alignment.} In the chemistry domain, a specialized advantage demonstrated by GPT-OSS is \textsc{\#1Reconciling Representations to Outcomes}. This comprehensive ability requires the model to seamlessly fuse multiple diverse descriptions of the same system. By jointly processing 1D symbolic stoichiometry, 2D topological connectivity, and dynamic state variables, the model effectively synthesizes these different chemical perspectives. By successfully organizing these features, the larger model can dynamically track complex chemical transformations and accurately project intermediate states into the correct final outcome.

\subsection{Validation via SLM Systemic Failures}

\begin{wrapfigure}{r}{0.5\textwidth}
  \centering
  \includegraphics[width=\linewidth]{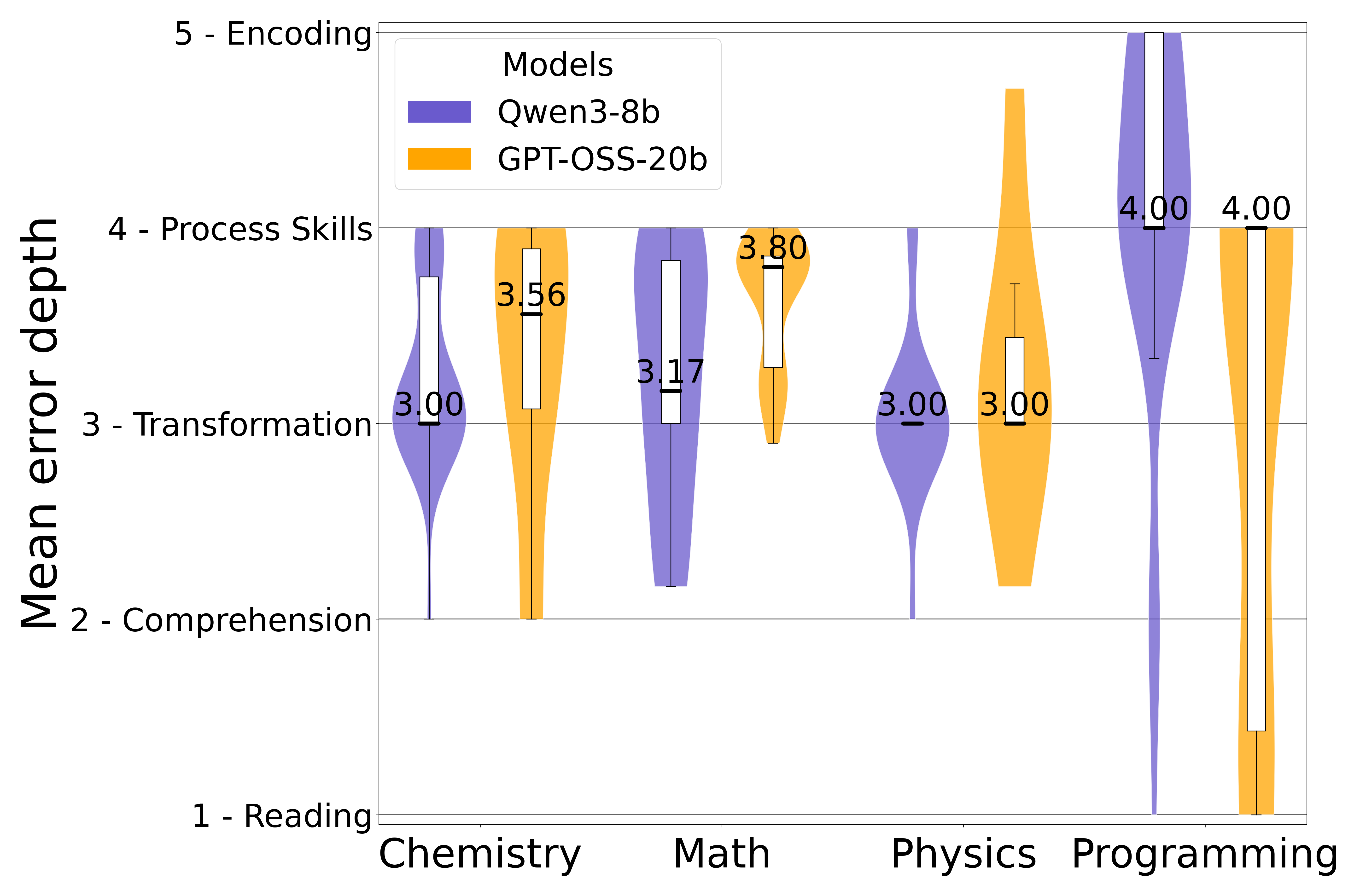}
  \caption{\textbf{Distribution of First Error Depths for Smaller Models.} The y-axis indicates the mean first-error depth across runs of a question, from 1 (Reading) to 5 (Encoding). Violin width represents failure density, with annotated horizontal lines marking the medians. Only consistent errors ($E_{norm} \le 0.4$) are included. Across all domains (Chemistry, Math, Physics, Programming), failures for smaller models heavily concentrate at 3- Transformation) and 4- Process Skills).}
  \label{fig:slm_errors}
\end{wrapfigure}

Beyond identifying the advantages of larger models, we validate our findings by analyzing where smaller models systematically fail. Using the Error Dispersion Index (EDI, Appendix ~\ref{sec:appendix_first_error}) to quantify "First-Error" consistency, we found that both Qwen3-8B and GPT-OSS-20B yield an average EDI $< 0.4$. This confirms that their failures are not stochastic, but rather stem from fundamental capability deficits.

As shown in Figure~\ref{fig:slm_errors}, these deficits heavily concentrate at the \textit{Transformation} (Depth 3) and \textit{Process Skills} (Depth 4) stages. This distribution directly mirrors the core advantage of larger models: Constraint-Guided Reasoning. Lacking the consistent support of this capability, smaller models either fail to abstract problem into bounded mathematical representations (Depth 3), or they resort to unconstrained, localized trial-and-error that inevitably diverges during execution (Depth 4). Ultimately, these bottlenecks provide empirical evidence that Constraint-Guided Reasoning constitutes the primary capability gap between smaller and larger models.
\section{Conclusion}
This work aims to better understand how reasoning changes with scale within the same model family. We develop AdvCluster, an automated analysis framework that systematically compares larger and smaller model reasoning traces, extracts  advantage descriptions, and organizes them through semantic clustering. This allows recurring patterns to emerge naturally from the data and yields a structured taxonomy of common advantages and specialized advantages. 

Our main finding is that larger models across different model families consistently exhibit Constraint-Guided Reasoning: they are better at identifying conditions, organizing them into structured constraints, and using them to guide reasoning and eliminate infeasible paths. We hope this work provides a clearer starting point for understanding the reasoning advantages that emerge with scale.
\clearpage

\bibliography{colm2026_conference}
\bibliographystyle{colm2026_conference}

\clearpage
\appendix
\section{Appendix}
\subsection{LLM Usage Summary}
\begin{table}[h]
    \centering
    \label{tab:llm_usage_summary}
    
    \begin{tabularx}{\textwidth}{@{} l l l >{\RaggedRight\arraybackslash}X @{}}
        \toprule
        \textbf{Pipeline Stage} & \textbf{Role} & \textbf{Model} & \textbf{Purpose} \\
        \midrule
        
        Advantage Extraction & Advantage extractor & Gemini 3 Pro & Compare larger and smaller model reasoning traces and generate advantage descriptions. \\
        \addlinespace[1ex]
        
        Semantic Clustering & Summarization model & GPT-5.2 & Summarize advantges descriptions and generate cluster tags and definitions. \\
        \addlinespace[1ex]
        
        Semantic Clustering & Reviewer model & GPT-5.2 & Review candidates in terms of semantic quality and select final clustering solution. \\
        \bottomrule
    \end{tabularx}
\end{table}
\clearpage

\subsection{Prompt}
\newtcolorbox{MyPromptBox}[1][]{
    colback=gray!20!white,   
    colframe=gray!50!black, 
    title=Prompt, 
    fonttitle = \bfseries\sffamily,
    boxrule=0.2pt
}
\subsubsection{Advantage Extractor prompt}

\begin{MyPromptBox}

You are an LLM advantage extraction expert. Model\_A is correct; Model\_B is incorrect.\\

TASK: \\
Compare their reasoning and do the following:\\
1. Identify the FIRST Newman's Error Analysis stage where Model\_B fails.\\
Choose ONE: Reading \textbar  Comprehension  \textbar  Transformation  \textbar  Process Skills  \textbar  Encoding \\
2. Briefly describe the specific reasoning failure at that stage.\\
3. From this failure, extract 2-5 advantage objects explaining why Model\_A succeeds.\\

!!! RULES FOR 'advantage' FIELD !!! \\
1. ABSTRACTION: Generalize to universal reasoning skills (no problem-specific variables). \\
2. FORMAT: Start EXACTLY with an action verb (e.g., "Identifies", "Applies"). \\
3. FORBIDDEN: NEVER use "Model\_A", "Model\_B", "The model", or "Correctly". \\
4. LENGTH: Maximum 10 words.\\

Problem: \{q\} \\
Model\_A reasoning: \{larger\_model\_reasoning\} \\
Model\_B reasoning: \{smaller\_model\_reasoning\} \\

Output Format:\\
Each object MUST follow this exact schema:\\
** ONLY OUTPUT LIST OF OBJECTS, NO OTHER TEXT. **
\begin{verbatim}
[
  {
    "type": "failure", 
    "failure_stage": "<ONE OF: Reading | Comprehension | Transformation | 
       Process Skills | Encoding>",
    "failure_description": "<Brief description of Model_B's FIRST failure>"
  },
  {
    "type": "advantage",
    "advantage": "<Action verb + abstract skill (max 10 words)>",
    "evidence": "<Specific text contrast proving the advantage>"
  },
  {
    "type": "advantage",
    "advantage": "<another Action verb + abstract skill (max 10 words)>",
    "evidence": "<another specific text contrast proving the advantage>"
  },...
]
\end{verbatim}
\end{MyPromptBox}
\label{llmanalyzeprompxt}
\clearpage

\subsubsection{Summarization model prompt}
\label{llmsummrization}
\begin{MyPromptBox}[breakable]
You are an expert in LLM Reasoning Analysis.
I will provide a cluster of similar reasoning advantages.\\
Please summarize the advantages and provide a tag and definition for the cluster.\\

INSTRUCTIONS:\\
1. DE-DOMAIN: Strip all subject-specific context (e.g., replace 'chemical valency' with 'structural constraints'). \\
2. ACTION-VERB TAGS: The tag MUST start with a strong action verb (e.g., Mapping, Verifying, Reducing, Deriving, Resolving). Avoid generic tags like 'Reasoning' or 'Logic'.\\
3. DEFINITION: Describe *how* the model manipulates information to reach a conclusion. Use 2-4 clear sentences.\\
4. LENGTH: Tag (2–4 words).\\

Advantages (bullet list):\\
\{advantage\_descriptions\_within\_cluster\_k\}\\

RETURN JSON ONLY in this format:
\begin{verbatim}
{ 
  "tag": "...",
  "definition": "..."
}
\end{verbatim}
\end{MyPromptBox}
\clearpage

\subsubsection{Reviewer Model prompt}
\label{llmreviewer}
\begin{MyPromptBox}[breakable]
You are evaluating a taxonomy of reasoning advantages induced from clustering.
Do not rewrite the taxonomy. Do not assume that fewer clusters are always better. Focus only on taxonomy quality. You will be given K and, for each cluster, an id, a tag, and a short definition.Evaluate the taxonomy using these criteria: \\

\#\#\# 1. Mutual Exclusivity (Distinctness)\\
- [1-2] Redundant: Two or more clusters describe the same concept.\\
- [3-4] Overlap: Significant overlap (\>40
- [5-6] Moderate: Different concepts, but boundaries are fuzzy/lack exclusion.\\
- [7-8] Sharp: Distinct features per cluster; very low ambiguity.\\
- [9-10] Exclusive: Logically impossible for an instance to belong to two clusters.\\

\#\#\# 2. Conceptual Precision \& Depth (Granularity)\\
- [1-2] Vague/Circular: Definitions offer no real insight or repeat the tag.\\
- [3-4] Surface-level: Describes ONLY the outcome (e.g., "the answer is wrong").\\
- [5-6] Functional: Describes the reasoning process/behavior in plain language.\\
- [7-8]Academic: Explains the underlying mechanism using formal logic terminology.\\
- [9-10] Research-Ready: High-level synthesis of unique reasoning failures/patterns; publication quality.\\

\#\#\# 3. Interpretability (Narrative Value)
- [1-2] Obscure: Hard to explain what kind of cases fit here.\\
- [3-4] Complex: Requires heavy mental effort to map to actual model behavior.\\
- [5-6] Clear: Understandable, but lacks a strong, cohesive "story."\\
- [7-8] Intuitive: A researcher can immediately picture the failure mode.\\
- [9-10] Insightful: Provides an "Aha!" moment; memorable and easy to communicate.\\

\#\#\# 4. Cluster Balance \& Utility (Granularity)\\
- [1-2] Failed: One dominant cluster (\>80\%) or useless fragmented noise.\\
- [3-4] Poor: Highly skewed\\
- [5-6] Passable: Distribution allows for basic statistical observation.\\
- [7-8] Healthy: Balanced enough to reveal clear performance trends across models.\\
- [9-10] Optimal: Ideal for high-precision comparative analysis and scaling law research.\\

\#\#\# 5. Taxonomy Resolution (Granularity) \\
- [1-3] Coarse: Too broad; merges distinct behaviors, losing diagnostic power.\\
- [4-6] Fragmented: Too noisy; over-splits clusters based on surface-level wording.\\
- [7-10] Functional: Just the right number of clusters to cover the main reasoning stages.\\

Input:
\begin{verbatim}
K = {k}
CLUSTERS = {clusters_section}    
\end{verbatim}
Output one JSON object only.

\begin{verbatim}
{{
  "Distinctness_score": [1-10],
  "Granularity_score": [1-10],
  "Interpretability_score": [1-10],
  "Balance_score": [1-10],
  "Taxonomy_resolution_score": [1-10],
}}
\end{verbatim}
\end{MyPromptBox}

\subsection{Final Clustering Solutions}
\label{sec: Final Clustering Solutions}
Tables~2 and~3 present the tags and definitions for all clusters in the final clustering solutions (Qwen3: $d'=8$, $K=6$; GPT-OSS: $d'=4$, $K=6$). These were generated during the semantic clustering stage and are included for reference. Table~4 reports the size of each final cluster.
\subsubsection{Tags and Definitions}
\begin{table}[H]
\begin{flushleft}
\small
\begin{tabularx}{1.1\textwidth}{@{} c >{\RaggedRight}p{1.6cm} X @{}}
\toprule
\textbf{Cluster ID} & \textbf{Tag} & \textbf{Definition} \\ \midrule
1 & Reconciling Representations to Outcomes & The model translates between multiple descriptions of the same system (symbolic counts, structural connectivity, state variables, and procedural conditions) and aligns them into a single coherent internal representation. It applies conservation-style constraints and rule-based mappings to infer implied structure, classify the transformation type, and predict resulting properties or end states. It then cross-checks consistency across sequential steps by tracking entities and quantities, rejecting candidate interpretations that violate constraints or fail to match observed indicators.
 \\ \addlinespace[1ex]

3 & Enforcing Constraint Consistency & The model extracts implicit and explicit constraints, then normalizes them into a unified set of structural rules, bounds, and admissibility conditions. It decomposes the space into disjoint cases/components, tracks invariants or monotone measures to prune candidates, and validates hypotheses by forward simulation on representative instances (including minimal/base cases) and by constructing witnesses for tightness. It repeatedly cross-checks intermediate results against all constraints—including edge cases and output-format requirements—to eliminate contradictory, loose, or underdetermined conclusions. \\ \addlinespace[1ex]

4 & Preserving Output Fidelity & Accurately transforms structured inputs into exact outputs by performing precise substitutions and calculations while keeping values exact and consistent. Maintains character- and element-level sequence integrity during parsing, indexing, counting, and concatenation, avoiding insertion, deletion, or hallucinated separators/whitespace. Enforces strict serialization and syntax/format constraints (e.g., option encoding, delimiter placement, length checks) and validates outputs via cross-checks against expected counts and components. Detects and corrects minor transcription errors (e.g., typos) by reconciling anomalies with global consistency and intended patterns.
 \\ \addlinespace[1ex]

5 & Deriving Tight Bounds & Transforms the task into an abstract constraint-and-counting model (e.g., inequalities, state graphs, dual programs), then propagates local requirements to global conclusions to force either feasibility or contradiction. It combines analytic bounding tools (counting, density/duality, pigeonhole/parity-style invariants, and recognized extremal theorems) with explicit extremal constructions that witness achievability or produce counterexamples. This approach targets tight upper/lower limits without exhaustive search by systematically reducing the space of possibilities and proving remaining cases impossible or realizable. \\ \addlinespace[1ex]

0 & Deriving Modular Periods & Transforms complex or unbounded processes into finite-state analyses by rewriting constraints as congruences and tracking state transitions modulo a chosen base. Identifies invariants and repeated residues to detect cycles, then computes the exact period using tools like multiplicative order and decomposition across independent factors. Generalizes the discovered periodic behavior into explicit formulas or closed-form recurrences that enable exact computation and impossibility proofs. \\ \addlinespace[1ex]

2 & Deriving Invariant Parameterizations & The model identifies quantities and relations that remain unchanged under allowable transformations, then uses them to re-encode the problem in a simpler coordinate/parameter space. It converts structural constraints into explicit equations or inequalities by applying symmetry arguments, variable substitutions, and algebraic normal forms (e.g., sum-of-squares, determinant/rank conditions, piecewise sign cases). From this reduced representation, it derives closed-form expressions, exact ratios/constants, and tight bounds, and verifies the result by back-substitution against the original constraints.
 \\ \bottomrule
\end{tabularx}
\end{flushleft}
\caption{GPT-OSS-120b Advantages Cluster Tags and Definition}
\end{table}

\begin{table}[H]
\centering
\small 
\begin{tabularx}{\textwidth}{@{} c >{\RaggedRight}p{2cm} X @{}}
\toprule
\textbf{Cluster ID} & \textbf{Tag} & \textbf{Definition} \\ \midrule
0 & Constraining via Invariants& The model translates varied representations into a common formal description, then computes derived quantities (e.g., counts, balances, configuration states, conserved totals) that must remain consistent across steps. It uses these invariants to eliminate candidates that violate quantitative or structural constraints, and to verify proposed conclusions by re-checking consistency from multiple angles. When general heuristics conflict with hard constraints or known exceptions, it prioritizes the constraint/exception and updates the conclusion accordingly.                 \\ \addlinespace[1ex]
2          & Recasting via Transformations     & The model systematically rewrites a problem into an equivalent representation where constraints and objectives become easier to enforce (e.g., shifting/aligning reference frames, converting aggregates into centered differences, or turning sums into products via alternate encodings). It tracks symbolic parameters through these transformations, preserving exact factors and required correction terms (such as scaling, adjustment multipliers, or change-of-variables weights) so the new form remains mathematically consistent. It then decomposes the transformed structure into components or phases (e.g., along principal directions or sequential regimes) to compute exact ratios, coefficients, and extrema under dependencies. \\ \addlinespace[1ex]
5          & Reducing to Existence Constraints & The model first recognizes the task as a yes/no existence decision, then rewrites the original requirements into a set of simpler equivalent constraints (e.g., divisibility-like conditions, linear constraints, or combinatorial counting conditions). It then exploits available degrees of freedom by selecting parameters sequentially so the constraints can be satisfied simultaneously, using either an explicit construction (assembling compatible local conditions into a global one) or a non-constructive argument (counting, contradiction, or contrapositive) to conclude existence or impossibility. Finally, it sanity-checks feasibility by testing boundary cases and verifying internal consistency of the derived constraints. \\ \addlinespace[1ex]
4          & Simulating Stateful Execution     & The model performs a literal step-by-step trace of a procedure by explicitly storing the current state (values, mappings, sequence contents) and updating it after every operation. It recomputes dependent quantities (e.g., relative positions, termination checks, branch conditions) against the updated state rather than the initial one, preserving order and handling shifting references caused by in-place modifications. It adheres strictly to stated control flow—even if it seems counterintuitive or buggy—and cross-checks the final result against intermediate states and invariant relationships to ensure consistency and completeness.               \\ \addlinespace[1ex]
1 & Verifying Exact Sequences         & Systematically simulates step-by-step sequence transformations at the element level, tracking indices, boundaries, and membership conditions to determine exactly which elements are kept, removed, or modified. Computes and cross-checks derived invariants (e.g., total length, adjacency, delimiter placement) against the constructed output to prevent omissions, insertions, or reordering. Distinguishes raw values from their encoded representations by applying the precise formatting/escaping rules required for valid literal output.\\ \addlinespace[1ex]
3 & Enforcing Output Conformance & Determines the task’s expected return form by identifying the question type and extracting explicit/implicit constraints on structure, type, and syntax. It then transforms derived results into fully resolved, concrete representations (not symbolic placeholders), aggregating all valid options into a single permissible output when multiple answers are allowed. Throughout, it preserves exact input values and verifies consistency by checking indexing, global structural properties, and signature/parameter requirements so the final text matches the required data object and formatting rules.   \\ \addlinespace[1ex] \bottomrule
\end{tabularx}
\caption{Qwen3-32b Advantages Cluster Tags and Definition}
\end{table}
\clearpage

\subsubsection{Cluster Sizes}
\begin{table}[H]
\centering
\begin{tabular}{lcc}
\toprule
\textbf{Model Family} & \textbf{Cluster ID} & \textbf{\# Advantage Descriptions} \\
\midrule
\multirow{6}{*}{Qwen3}
& 0 & 350 \\
& 1 & 246 \\
& 2 & 331 \\
& 3 & 233 \\
& 4 & 217 \\
& 5 & 447 \\
\midrule
\multirow{6}{*}{GPT-OSS}
& 0 & 316 \\
& 1 & 404 \\
& 2 & 332 \\
& 3 & 304 \\
& 4 & 184 \\
& 5 & 423 \\
\bottomrule
\end{tabular}
\caption{Number of advantage descriptions assigned to each cluster in the final clustering solutions.}
\label{tab:cluster_size_by_family}
\end{table}
\clearpage

\subsection{Supplementary Data}
\label{detailed_acc}
\subsubsection{Accuracy on Benchmarks}
The following four tables contain detailed data from the preliminary results across all datasets, with each table corresponding to one of the four domains: mathematical, physics, chemistry, and programming.

\begin{table}[H]
    \centering
    \renewcommand{\arraystretch}{1.4}
    \begin{tabular}{l ccc | c}
        \toprule
        \textbf{Model} & \textbf{HHMT} & \textbf{JEEBENCH} & \textbf{OMNI-MATH} & \textbf{Mean} \\
        \midrule
        Qwen3-8b     & $55.67 \pm 4.98$ & $92.75 \pm 1.27$ & $44.56 \pm 1.17$ & $64.33 \pm 2.47$  \\
        Qwen3-32b    & $\bm{60.00 \pm 3.14}$ & $\bm{95.64 \pm 0.53}$ & $\bm{49.65 \pm 0.92}$ & $\bm{68.43 \pm 1.53}$ \\
        GPT-OSS-20b  & $71.00 \pm 6.10$ & $83.35 \pm 1.13$ & $32.01 \pm 1.58$ & $62.12 \pm 2.94$ \\
        GPT-OSS-120b & $\bm{85.33 \pm 2.81}$ & $\bm{84.19 \pm 0.63}$ & $\bm{43.96 \pm 0.99}$ & $\bm{71.16 \pm 1.48}$ \\
        \bottomrule
    \end{tabular}
    \caption{Accuracy (\%) of various models on mathematical reasoning benchmarks}
    \label{tab:results_math}
\end{table}

\vspace{1.5em} 

\begin{table}[H]
    \centering
    \renewcommand{\arraystretch}{1.4}
    \begin{tabular}{l ccc | c}
        \toprule
        \textbf{Model} & \textbf{JEEBENCH} & \textbf{OlympiadBench} & \textbf{GPQA} & \textbf{Mean} \\
        \midrule
        Qwen3-8b     & $81.14 \pm 2.84$ & $26.37 \pm 44.06$ & $59.14 \pm 1.61$ & $55.55 \pm 16.17$ \\
        Qwen3-32b    & $\bm{88.37 \pm 1.44}$ & $\bm{31.33 \pm 46.38}$ & $\bm{69.09 \pm 2.11}$ & $\bf{62.93 \pm 16.64}$ \\
        GPT-OSS-20b  & $67.48 \pm 2.27$ & $\bm{29.56 \pm 1.97}$ & $91.63 \pm 2.04$ & $62.89 \pm 2.09$ \\
        GPT-OSS-120b & $\bm{70.16 \pm 1.84}$ & $29.29 \pm 2.45$ & $\bm{94.53 \pm 2.05}$ & $\bm{64.66 \pm 2.11}$ \\
        \bottomrule
    \end{tabular}
    \caption{Accuracy (\%) of various models on physics reasoning benchmarks}
    \label{tab:results_physics}
\end{table}

\vspace{1.5em}

\begin{table}[H]
    \centering
    \renewcommand{\arraystretch}{1.4}
    \begin{tabular}{l cc | c}
        \toprule
        \textbf{Model} & \textbf{JEEBENCH} & \textbf{GPQA} & \textbf{Mean} \\
        \midrule
        Qwen3-8b     & $65.19 \pm 2.22$ & $59.14 \pm 2.05$ & $62.17 \pm 2.14$ \\
        Qwen3-32b    & $\bm{78.27 \pm 1.43}$ & $\bm{68.03 \pm 1.93}$ & $\bm{73.15 \pm 1.68}$ \\
        GPT-OSS-20b  & $57.56 \pm 2.61$ & $43.66 \pm 2.83$ & $50.61 \pm 2.72$ \\
        GPT-OSS-120b & $\bm{70.06 \pm 1.98}$ & $\bm{65.91 \pm 3.44}$ & $\bm{67.99 \pm 2.71}$ \\
        \bottomrule
    \end{tabular}
    \caption{Accuracy (\%) of various models on chemistry reasoning benchmarks}
    \label{tab:results_chemistry}
\end{table}

\vspace{1.5em}

\begin{table}[H]
    \centering
    \renewcommand{\arraystretch}{1.4}
    \begin{tabular}{l cc | c}
        \toprule
        \textbf{Model} & \textbf{CRUXEVAL-I} & \textbf{CRUXEVAL-O} & \textbf{Mean} \\
        \midrule
        Qwen3-8b     & $22.46 \pm 0.70$ & $88.21 \pm 0.43$ & $55.34 \pm 0.57$ \\
        Qwen3-32b    & $\bm{25.23 \pm 0.57}$ & $\bm{92.00 \pm 0.74}$ & $\bm{58.61 \pm 0.66}$ \\
        GPT-OSS-20b  & $29.36 \pm 0.58$ & $95.50 \pm 0.74$ & $62.43 \pm 0.66$ \\
        GPT-OSS-120b & $\bm{30.16 \pm 0.67}$ & $\bm{97.31 \pm 0.60}$ & $\bm{63.74 \pm 0.63}$ \\
        \bottomrule
    \end{tabular}
    \caption{Accuracy (\%) of  models on programming reasoning benchmarks}
    \label{tab:results_programming}
\end{table}

\subsubsection{Analysis Question Set}
\label{Analysis Question Set by Subject}
Table~9 provides the detailed counts of the 115 (Qwen3) and 106 (GPT-OSS) questions retained after gap-based filtering, categorized by domain.
\begin{table}[H]
\centering

\begin{tabular}{llcc}
\toprule
\textbf{Model Family} & \textbf{Domain} & \textbf{Retained Questions} & \textbf{Total Questions in Pool} \\
\midrule
\multirow{5}{*}{Qwen3}
& Math        & 21 & 748 \\
& Physics     & 16 & 322  \\
& Chemistry   & 27 & 249  \\
& Programming & 51 & 1600 \\
& Total       & 115 & -- \\
\midrule
\multirow{5}{*}{GPT-OSS}
& Math        & 53 & 748  \\
& Physics     & 8  & 322  \\
& Chemistry   & 27 & 249  \\
& Programming & 18 & 1600 \\
& Total       & 106 & -- \\
\bottomrule
\end{tabular}
\caption{Number of analysis questions retained after the gap-based filter, by domains.}
\label{tab:analysis_question_set_by_subject}
\end{table}

\subsubsection{Advantage Descriptions  Statistics}
Table~\ref{tab:dedup_stats} summarizes the effect of semantic deduplication on the extracted advantage descriptions for each model pair.
\label{Advantage Descriptions  Statistics}

\begin{table}[H]
\centering
\begin{tabular}{lcccc}
\toprule
\textbf{Model } & \textbf{Before Dedup} & \textbf{After Dedup} & \textbf{Removed} & \textbf{Removal Ratio (\%)} \\
\midrule
Qwen3-32b vs.\ Qwen3-8b         & 1927 & 1824 & 103 & 5.34 \\
GPT-OSS-120b vs.\ GPT-OSS-20b   & 2019 & 1963 & 56  & 2.77 \\
\bottomrule
\end{tabular}
\caption{Deduplication statistics for extracted advantage descriptions.}
\label{tab:dedup_stats}
\end{table}
\clearpage

\subsection{Experimental Details for First-Error Analysis}
\label{sec:appendix_first_error}

To investigate the systemic bottlenecks of smaller models, we conducted the "First-Error" analysis. Our goal is to determine whether these models make stochastic (random) mistakes or exhibit recurring error.

\paragraph{First-Error Extraction Pipeline}
We filter for challenging questions for smaller models, where the performance gap between the larger and smaller models is $\ge 0.6$. For each question, we prompt the Advantage Extractor to identify the exact reasoning step where the smaller model commits its First Error during the Advantage Extraction stage. We adapt Newman's Error Analysis framework, grouping reasoning failures into five sequential depths (1 to 5):
\begin{enumerate}[leftmargin=20pt, itemsep=2pt, parsep=0pt]
    \item \textbf{Reading:} Failing to recognize basic text, symbols, or vocabulary.
    \item \textbf{Comprehension:} Misunderstanding the overall objective or meaning of the problem.
    \item \textbf{Transformation:} Failing to translate the problem description into a valid mathematical or logical operation.
    \item \textbf{Process Skills:} Committing errors during step-by-step calculations or rule execution.
    \item \textbf{Encoding:} Failing to properly format or express the final derived solution.
\end{enumerate}

\paragraph{Error Dispersion Index (EDI) Formulation}
To quantify error consistency, we propose the Error Dispersion Index (EDI), a question-level metric based on normalized empirical Shannon entropy. For a given question, assuming $N$ independent reasoning trajectories sampled from a smaller model, let $c_i$ be the count of "first errors" occurring at depth $i$ out of $|D|=5$ possible depths. The empirical probability is $P(d_i) = \frac{c_i}{N}$.

The empirical Shannon entropy (in bits) for this specific question is calculated as:
\begin{equation}
    H = - \sum_{P(d_i)>0} P(d_i) \log_2(P(d_i))
\end{equation}

To ensure the metric is bounded within $[0, 1]$, we normalize $H$ by the exact empirical maximum entropy, $\mathcal{H}_{max}(N, |D|)$. Conceptually representing a state of completely random guessing, $\mathcal{H}_{max}$ is defined as the entropy achieved when the $N$ discrete errors are distributed as evenly as possible across the $|D|$ reasoning depths. The per-question EDI is then defined as:
\begin{equation}
    EDI = \frac{H}{\mathcal{H}_{max}(N, |D|)}
\end{equation}

\begin{itemize}[leftmargin=20pt, itemsep=2pt, parsep=0pt]
    \item $EDI \to 1$: Errors are uniformly scattered (stochastic).
    \item $EDI \to 0$: Errors isolate to a single depth (deterministic).
\end{itemize}
\clearpage

\subsection{Summary of Error Dispersion Data}
While the EDI is computed independently for each question, we average these scores across all questions within a specific domain to evaluate broader systemic trends. As detailed in Table~\ref{tab:edi_results}, across all domains, smaller models consistently yield an average EDI below $0.4$. This confirms that their reasoning failures are predominantly structural deficits rather than stochastic hallucinations.

\begin{table}[htbp]
\centering
\begin{tabular}{llccc}
\toprule
\textbf{Domain} & \textbf{Model} & $\mathbf{N_Q}$ & \textbf{Mean EDI} & \textbf{Std EDI} \\
\midrule
\multirow{2}{*}{Chemistry}   & GPT-OSS-20B & 12 & 0.136 & 0.148 \\
                             & Qwen3-8B    & 29 & 0.114 & 0.151 \\
\midrule
\multirow{2}{*}{Math}        & GPT-OSS-20B & 29 & 0.272 & 0.109 \\
                             & Qwen3-8B    & 13 & 0.222 & 0.129 \\
\midrule
\multirow{2}{*}{Physics}     & GPT-OSS-20B & 7  & 0.175 & 0.168 \\
                             & Qwen3-8B    & 17 & 0.052 & 0.119 \\
\midrule
\multirow{2}{*}{Programming} & GPT-OSS-20B & 9  & 0.054 & 0.108 \\
                             & Qwen3-8B    & 35 & 0.054 & 0.111 \\
\bottomrule
\end{tabular}
\caption{Summary of Error Dispersion Index (EDI) across domains and lower-capacity models. $N_Q$ denotes the number of challenging questions analyzed. A lower Mean EDI indicates that the model's first-errors are highly concentrated at specific reasoning depths.}
\label{tab:edi_results}
\end{table}

\end{document}